\documentclass[conference]{IEEEtran}
\IEEEoverridecommandlockouts
\usepackage{latex8}
\usepackage{amsmath,amssymb,amsfonts}
\usepackage{algorithmic}
\usepackage{url}
\usepackage{cite}
\usepackage{graphicx}
\usepackage{textcomp}
\usepackage{xcolor}
\usepackage{multirow}
\usepackage{caption}
\usepackage[shortlabels]{enumitem}
\usepackage[colorlinks=true, linkcolor=blue, urlcolor=blue, citecolor=blue, breaklinks=true]{hyperref}
\begin{document}

\title{ Degradation-Aware Image Enhancement via Vision-Language Classification }

\author{
\large{ Jie Cai, Kangning Yang, Jiaming Ding, Lan Fu, Ling Ouyang, Jiang Li, Jinglin Shen, Zibo Meng }\\ \\
\normalsize{OPPO AI Center, Palo Alto, CA, US}\\
\normalsize{ jie.cai@oppo.com }\\
}

\maketitle

\begin{abstract}

Image degradation is a prevalent issue in various real-world applications, affecting visual quality and downstream processing tasks. In this study, we propose a novel framework that employs a Vision-Language Model (VLM) to automatically classify degraded images into predefined categories. The VLM categorizes an input image into one of four degradation types: (A) super-resolution degradation (including noise, blur, and JPEG compression), (B) reflection artifacts, (C) motion blur, or (D) no visible degradation (high-quality image). Once classified, images assigned to categories A, B, or C undergo targeted restoration using dedicated models tailored for each specific degradation type. The final output is a restored image with improved visual quality. Experimental results demonstrate the effectiveness of our approach in accurately classifying image degradations and enhancing image quality through specialized restoration models. Our method presents a scalable and automated solution for real-world image enhancement tasks, leveraging the capabilities of VLMs in conjunction with state-of-the-art restoration techniques.

\end{abstract}


\section{Introduction}
\label{sec:introduction}

Image degradation significantly impacts the performance of computer vision tasks and overall visual perception. Common degradation types, such as noise, blur, compression artifacts, reflection artifacts, and motion blur, often arise due to limitations in imaging conditions, sensor quality, and environmental factors. Traditional image restoration techniques typically require predefined knowledge about the type of degradation, limiting their adaptability in real-world scenarios.

To address this challenge, we propose an automated image degradation classification and restoration pipeline that utilizes a Vision-Language Model (VLM). The VLM processes an input consisting of an image and a textual prompt:
\begin{quote}
"Analyze this image and determine the type of image degradation it exhibits. Categorize it into one of the following degradation types: A. Super-resolution degradation (including noise, blur, JPEG compression); B. Reflection artifacts; C. Motion blur; D. No visible degradation (high-quality image). Provide a simple result, i.e. A, B, C, or D."
\end{quote}
Based on the classification output, the degraded image is passed through a corresponding specialized restoration model for enhancement.

For images classified under super-resolution degradation (category A), we utilize the InvSR model~\cite{yue2025arbitrary}, which is particularly effective for general super-resolution but struggles with text regions. To compensate for this limitation, we employ PaddleOCR~\cite{baidu2021paddleocr} to detect text areas and apply Real-ESRGAN~\cite{wang2021real} for targeted text restoration. The final result is obtained by fusing the InvSR-processed whole image with the Real-ESRGAN-enhanced text regions.
For images affected by reflection artifacts (category B), we first detect strong reflections using YOLO~\cite{wang2024yolov9} and YOSO~\cite{hu2023you}, generating a reflection mask. The masked regions are then inpainted using the LaMa model~\cite{suvorov2022resolution} to remove strong reflections. Finally, we apply NAFNet~\cite{chen2022simple} to further refine the image by reducing weak reflections.
For motion-blurred images (category C), we first employ a NAFNet-based deblurring model~\cite{chen2022simple} to restore general image sharpness. In cases where human faces are present, we further enhance facial details using CodeFormer~\cite{zhou2022towards}, ensuring high-quality facial reconstruction and refinement.

By combining VLM-based classification with specialized restoration models, our framework provides an automated and effective solution for image enhancement. Experimental results show that our method achieves high accuracy in degradation classification and significant improvements in visual quality across various degradation types. This approach offers a scalable and efficient way to enhance real-world images, bridging the gap between degradation classification and restoration.

\section{Related Work}
\label{sec:related_work}

\subsection{Vision-Language Models (VLMs)}
Vision-Language Models (VLMs)~\cite{liu2023visual,liu2024improved,li2022blip,alayrac2022flamingo,bai2025qwen2,zhu2025internvl3} have emerged as powerful tools for processing multimodal data by integrating visual and textual information. CLIP~\cite{radford2021learning} is a pioneering VLM that employs contrastive learning to align images and text in a shared embedding space, enabling strong zero-shot and few-shot learning capabilities. It has demonstrated remarkable performance in image classification, retrieval, and open-vocabulary tasks without task-specific fine-tuning.

Building on these advancements, Qwen2.5-VL~\cite{bai2025qwen2} further enhances vision-language understanding with improved visual recognition, precise object localization, and robust document parsing. As the leading open-source VLM, it excels in structured data extraction, long-video comprehension, and real-world interaction, making it highly adaptable for various applications. In the context of image restoration, VLMs like CLIP and Qwen2.5-VL can be leveraged to classify degradation types, serving as a critical first step in an automated restoration pipeline by providing high-level semantic understanding of degraded images.

\subsection{Image Super-Resolution}
Image super-resolution techniques~\cite{cai2022real,wang2021real,yu2024scaling,wu2024one,yue2025arbitrary} aim to enhance image quality by reconstructing high-resolution details from degraded low-resolution inputs. Deep learning-based SR models, such as Real-ESRGAN~\cite{wang2021real}, have significantly improved perceptual quality by introducing adversarial training and perceptual loss. Notably, Real-ESRGAN extends its capabilities to real-world restoration tasks. The model is trained exclusively on synthetic data, enabling robust and practical performance in diverse degradation scenarios.

SUPIR (Scaling-UP Image Restoration)~\cite{yu2024scaling} is a novel image restoration method that leverages generative priors, multi-modal techniques, and model scaling to achieve high-quality, realistic restoration. By training on a large-scale dataset with text annotations and introducing restoration-guided sampling and negative-quality prompts, SUPIR enables text-driven image restoration with enhanced perceptual fidelity.
OSEDiff~\cite{wu2024one} introduces a one-step diffusion network for real-world image super-resolution (Real-ISR), eliminating the need for multi-step diffusion by using the low-quality image as the starting point instead of random noise. By integrating variational score distillation for regularization, OSEDiff achieves high-quality restoration efficiently, outperforming existing diffusion-based Real-ISR methods in both accuracy and computational cost.
InvSR~\cite{yue2025arbitrary} leverages diffusion inversion with a Partial Noise Prediction strategy to initialize the sampling process at an intermediate diffusion state, improving image super-resolution efficiency. Its deep noise predictor enables flexible sampling steps (1-5), achieving state-of-the-art performance even with a single step.

Diffusion-based models have demonstrated superior performance over Real-ESRGAN in most image super-resolution tasks, particularly in generating high-quality textures and fine details. However, they struggle with text restoration, often producing artifacts or distorted characters. To address this limitation, we adopt a hybrid approach: we first apply text detection to identify regions containing text, then use Real-ESRGAN to restore these areas while leveraging the diffusion model for the rest of the image. This strategy effectively combines the strengths of both methods, ensuring high-quality restoration for both general content and text regions.

\subsection{Reflection Removal}
In academic research, most papers~\cite{yang2025ntire,yang2025openrr1k,yang2025survey,cai2025f2t2hit,cai2025openrr5k,hu2023single,zhu2024revisiting,zhao2025reversible} are evaluated on limited training and test datasets, which significantly restricts the capabilities of models. In real-world scenarios, reflections vary widely in both types and intensities, such as strong reflections, which are often beyond the scope of academic models. To address this, we design a strategy for detecting and inpainting strong reflections, which is then incorporated into a reflection separation network to more effectively handle diverse reflection scenarios.

This paper~\cite{hu2023single} introduces a generalized reflection superposition model with a learnable residue term to enhance decomposition completeness. By leveraging a dual-stream interaction mechanism and a semantic pyramid encoder, the proposed method achieves state-of-the-art performance across multiple real-world benchmarks.
This study~\cite{zhu2024revisiting} presents a large-scale reflection dataset, Reflection Removal in the Wild (RRW), and a novel Maximum Reflection Filter (MaxRF) to improve single-image reflection removal (SIRR) in real-world scenarios. By utilizing a reflection location-aware cascaded framework, the proposed method outperforms existing approaches on various benchmarks.
The Reversible Decoupling Network (RDNet)~\cite{zhao2025reversible} addresses limitations in existing reflection removal models by using a reversible encoder to preserve valuable information and flexibly decouple transmission and reflection features. With the addition of a transmission-rate-aware prompt generator, RDNet outperforms state-of-the-art methods on five widely-adopted benchmark datasets.

\subsection{Motion Deblurring}

NAFNet~\cite{chen2022simple} is a lightweight and efficient image restoration model that eliminates the need for traditional nonlinear activation functions, achieving state-of-the-art performance with significantly reduced computational costs. Given its effectiveness, we utilize NAFNet for image deblurring, leveraging its superior PSNR performance on benchmarks like GoPro while maintaining high efficiency.

CodeFormer~\cite{zhou2022towards} is a transformer-based blind face restoration model that leverages a learned discrete codebook prior to reduce ambiguity and enhance high-quality detail generation. We utilize CodeFormer for specialized face enhancement after applying NAFNet for overall deblurring, particularly addressing motion blur caused by human movement, such as in real-world scenarios involving children or dynamic subjects.

\section{Methodology}
\label{sec:methodology}

We propose a framework that combines Vision-Language Models (VLMs) with specialized restoration techniques for different image degradations. First, the VLM classifies the image into one of four degradation types: super-resolution degradation, reflection artifacts, motion blur, or no degradation. Based on this classification, we apply corresponding restoration models to enhance image quality. In this section, we describe the restoration methods used for each degradation type, including those for super-resolution, reflection artifacts, and motion blur.

\subsection{Vision-Language Model for Degradation Classification}

\begin{figure*}[ht]
    \centering
    \includegraphics[width=0.7\textwidth]{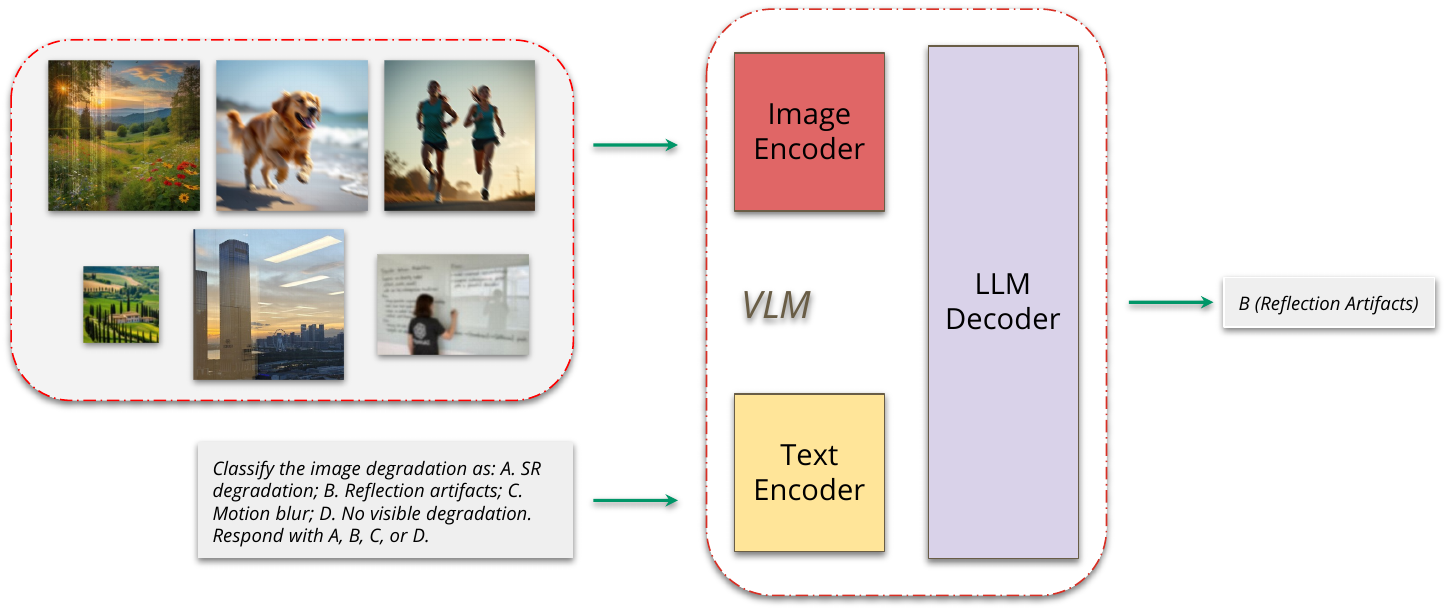}
    \caption{Vision-Language Model (VLM) architecture for zero-shot classification of image degradation types.}
    \label{fig:vlm_architecture}
\end{figure*}

As shown in Fig.~\ref{fig:vlm_architecture}, we utilize Qwen2.5-VL~\cite{bai2025qwen2} for degradation-aware vision-language classification, leveraging its advanced visual recognition and robust document parsing capabilities. As the leading open-source vision-language model in the industry, Qwen2.5-VL stands out for its superior performance in structured data extraction, object localization, and long-video comprehension. Its dynamic resolution processing and native ViT architecture make it particularly well-suited for handling degraded visual inputs, ensuring accurate classification even under challenging conditions. The openness and state-of-the-art capabilities of Qwen2.5-VL are the primary reasons for our selection, as it provides both cutting-edge performance and scalability for real-world applications.

Our approach leverages Vision-Language Models (VLMs) to classify image degradation types in a zero-shot manner. Instead of fine-tuning the VLM on specific image degradation datasets, we take advantage of the model's inherent zero-shot capabilities, enabling it to understand and classify the degradation types directly from the input image and the provided textual prompt. This approach allows us to apply a powerful pretrained VLM without the need for extensive retraining or task-specific data.

Given an input image, the VLM is fed with a textual prompt: 
\begin{quote}
"Analyze this image and determine the type of image degradation it exhibits. Categorize it into one of the following degradation types: A. Super-resolution degradation (including noise, blur, JPEG compression); B. Reflection artifacts; C. Motion blur; D. No visible degradation (high-quality image). Provide a simple result, i.e. A, B, C, or D."
\end{quote}
The VLM processes this information and classifies the image into one of the four predefined categories. This zero-shot classification task benefits from the model’s extensive pretraining on a large corpus of multimodal data, enabling accurate degradation identification without requiring any task-specific adjustments.
Once the degradation type is identified, the corresponding image restoration model is applied, which is tailored for the specific degradation category. This combination of VLM-based classification and specialized restoration models ensures that each type of degradation is handled effectively.

The framework enables a seamless and automated solution for image restoration, using the VLM’s zero-shot abilities to classify the image and guide the application of targeted restoration techniques.

\subsection{Image Super-Resolution}

\begin{figure*}[ht]
    \centering
    \includegraphics[width=0.7\textwidth]{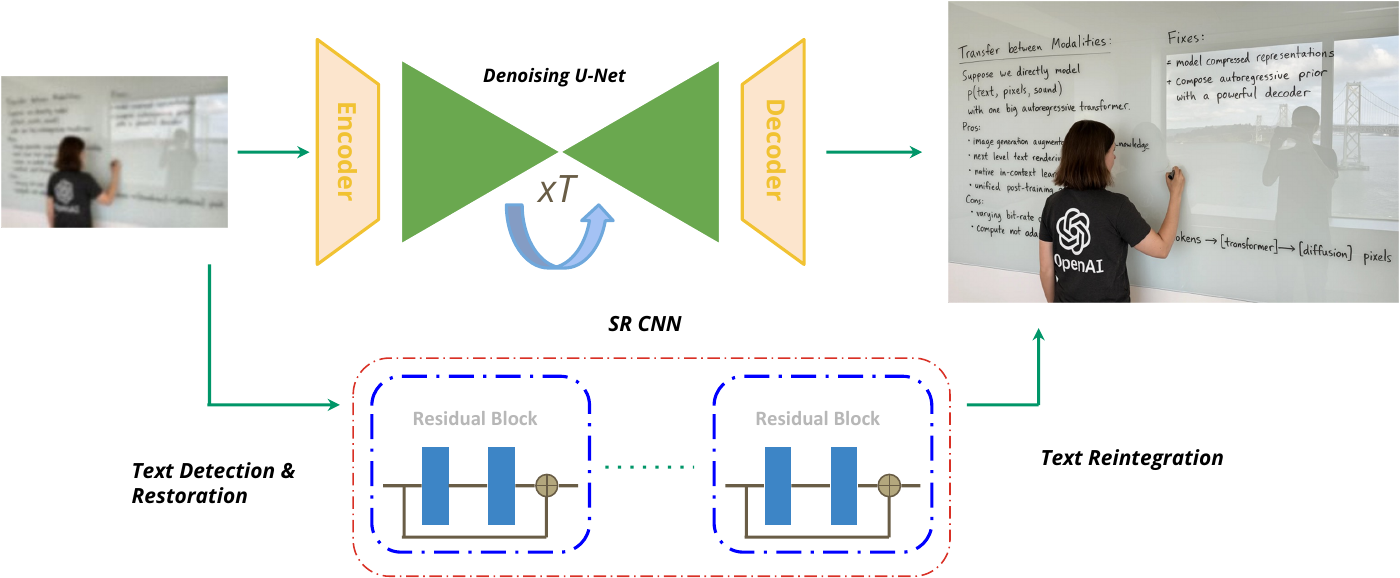}
    \caption{The architecture for InvSR super-resolution restoration, combined with text extraction and restoration for improved visual quality in text regions.}
    \label{fig:invsr_architecture}
\end{figure*}

For the restoration of images suffering from super-resolution degradation (including noise~\cite{cai2022real,cai2024joint}, blur, and JPEG compression), we employ the InvSR model~\cite{yue2025arbitrary}, which is an advanced image super-resolution technique based on diffusion inversion. The primary goal of InvSR is to leverage the rich priors encapsulated in large pre-trained diffusion models to enhance the resolution of images.

InvSR utilizes a novel Partial Noise Prediction strategy to construct an intermediate state in the diffusion process, which serves as the initial sampling point. This intermediate state is generated by a deep noise predictor that estimates the optimal noise maps for the forward diffusion process. Once the model is trained, this noise predictor facilitates the initialization of the sampling process at various stages of the diffusion trajectory, allowing the generation of a high-resolution image from a low-resolution input.
A key advantage of InvSR is its flexible and efficient sampling mechanism, which supports a wide range of sampling steps, from one to five. Even when using a single sampling step, InvSR achieves performance that is either superior or comparable to existing state-of-the-art super-resolution methods. This makes InvSR particularly effective in restoring high-quality images even with minimal computational resources.

For text-heavy images, InvSR's performance may degrade in the presence of severe compression or noise in textual regions. To address this, we complement InvSR~\cite{yue2025arbitrary} with PaddleOCR~ \cite{baidu2021paddleocr} to detect and separate text regions, followed by Real-ESRGAN~\cite{wang2021real} to specifically enhance these textual details. Finally, as shown in Fig.~\ref{fig:invsr_architecture}, the restored text regions are combined with the output from InvSR to produce the final high-resolution image.

\subsection{Reflection Removal}

\begin{figure*}[ht]
    \centering
    \includegraphics[width=0.7\textwidth]{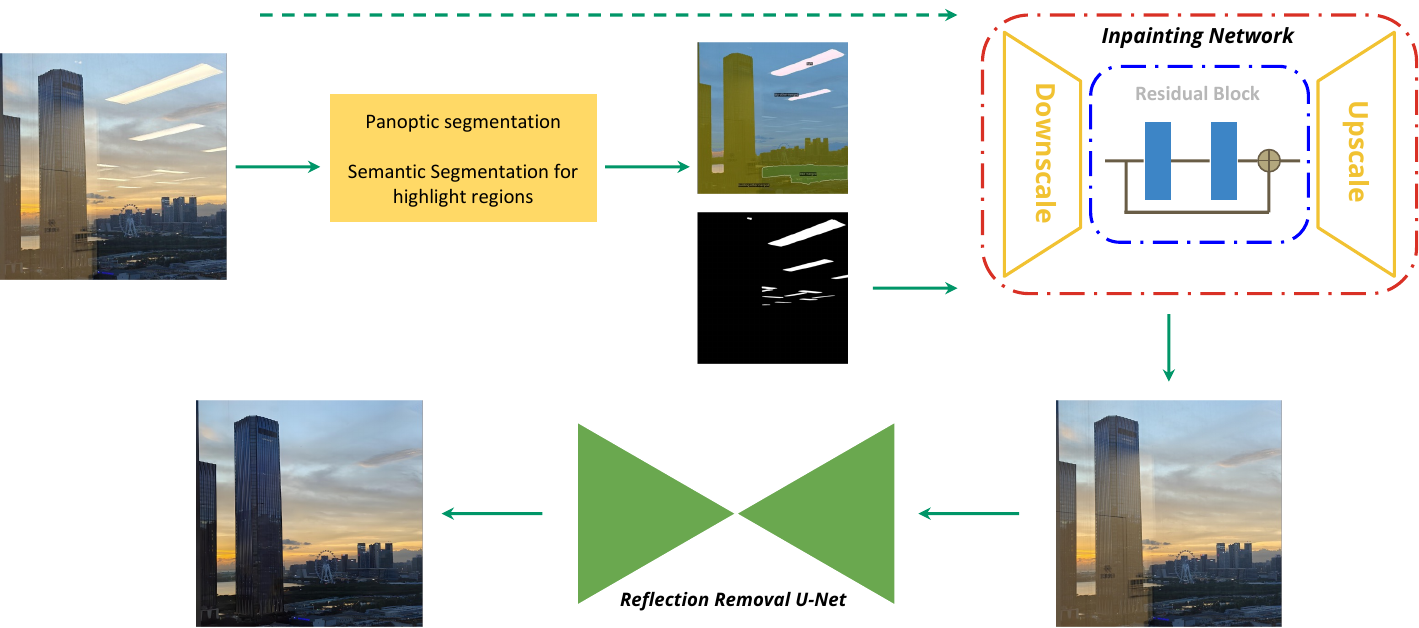}
    \caption{Reflection artifact removal architecture. It uses YOLO/YOSO for reflection detection and LaMa for inpainting the masked regions. Then, it used NAFNet for reflection removal.}
    \label{fig:reflection_removal_architecture}
\end{figure*}

For images affected by reflection artifacts (category B), we utilize a multi-step approach involving YOLO~\cite{wang2024yolov9}, YOSO~\cite{hu2023you}, LaMa~\cite{suvorov2022resolution}, and NAFNet~\cite{chen2022simple} models to detect and restore regions affected by strong and weak reflections, as shown in Fig.~\ref{fig:reflection_removal_architecture}.

First, we employ the YOLO object detection model to identify regions of the image containing strong reflection artifacts. YOLO is well-suited for this task due to its real-time performance and ability to accurately detect various object types, including reflections. Once strong reflections are detected, a reflection mask is created, which highlights the areas that need restoration. Next, we use YOSO, a segmentation model designed for segmentation tasks in images, to further refine the reflection mask. YOSO can segment the image at a pixel level, ensuring that we isolate only the regions affected by reflections, including smaller and less obvious reflection artifacts. This helps in accurately targeting the areas requiring restoration.
After generating a mask for the strong reflection regions, we apply LaMa~\cite{suvorov2022resolution}, an image inpainting model that is particularly effective for filling in missing or corrupted parts of an image. LaMa excels in reconstructing areas that have been affected by reflection artifacts, restoring the image to a more natural appearance. LaMa’s resolution-aware inpainting technique ensures that the restoration preserves fine details, especially around the edges of the reflection regions. To further enhance the image, we apply NAFNet~\cite{chen2022simple} to address any remaining weak reflection artifacts. NAFNet, with its nonlinear activation-free network, is highly effective in restoring subtle image details and reducing the visibility of weak reflections. This step refines the image by ensuring that both strong and weak reflections are adequately handled.

By combining these models, we effectively remove reflection artifacts, restoring the image to a visually pleasing state. The approach not only recovers strong reflections but also refines the image by handling weaker artifacts, ensuring a high-quality restoration for images affected by reflection degradation.

\subsection{Motion Blur Removal}

\begin{figure*}[ht]
    \centering
    \includegraphics[width=0.7\textwidth]{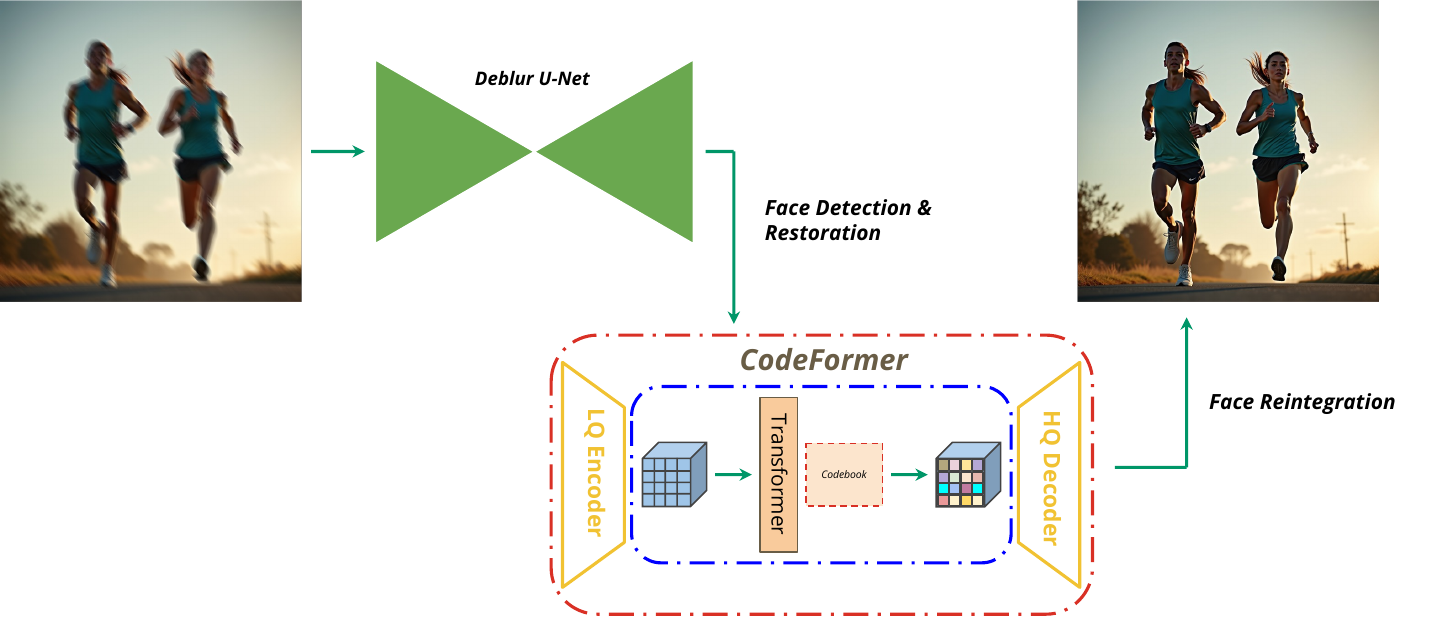}
    \caption{Motion blur restoration architecture, including NAFNet for deblurring and CodeFormer for facial enhancement.}
    \label{fig:motion_blur_architecture}
\end{figure*}

For images affected by motion blur (category C), we employ a two-step restoration process that utilizes the NAFNet~ \cite{chen2022simple} model for deblurring and the CodeFormer~\cite{zhou2022towards} model for further enhancing facial details, as shown in Fig.~\ref{fig:motion_blur_architecture}.

In the first step, we use NAFNet, a state-of-the-art image restoration network that is designed to handle various image degradation tasks, including deblurring. NAFNet utilizes a nonlinear activation-free architecture, which helps in reducing artifacts and improving image quality. It effectively restores sharpness in images that have been degraded by motion blur, recovering both fine details and overall image clarity. This makes NAFNet highly suitable for general motion blur removal, where traditional methods might fail to restore fine textures and sharpness in the image.
However, in cases where human faces are present in the image, additional restoration is required to ensure high-quality facial details. Motion blur often causes faces to lose their sharpness and clarity, making the restoration of facial features crucial for achieving a realistic result. To address this, we apply CodeFormer, a model specifically designed for robust face restoration with transformers. CodeFormer uses a transformer-based framework to refine facial features, restoring high-resolution facial details even in images suffering from severe motion blur. This model is particularly effective in maintaining the identity of individuals and enhancing facial features like eyes, nose, and mouth, ensuring that the restored image is both realistic and visually appealing.

By combining NAFNet for general deblurring and CodeFormer for specialized face restoration, we ensure that both the overall sharpness of the image and the fidelity of facial details are restored, providing a comprehensive solution for motion blur removal.

\section{Experiments}
\label{sec:experiments}

In this section, we present the visual results of our framework applied to three types of image degradation: super-resolution degradation, reflection artifacts, and motion blur. For each degradation type, we show three representative images, displaying the original degraded image and our restoration result. 

\begin{figure*}[ht]
    \centering
    \includegraphics[width=0.95\textwidth]{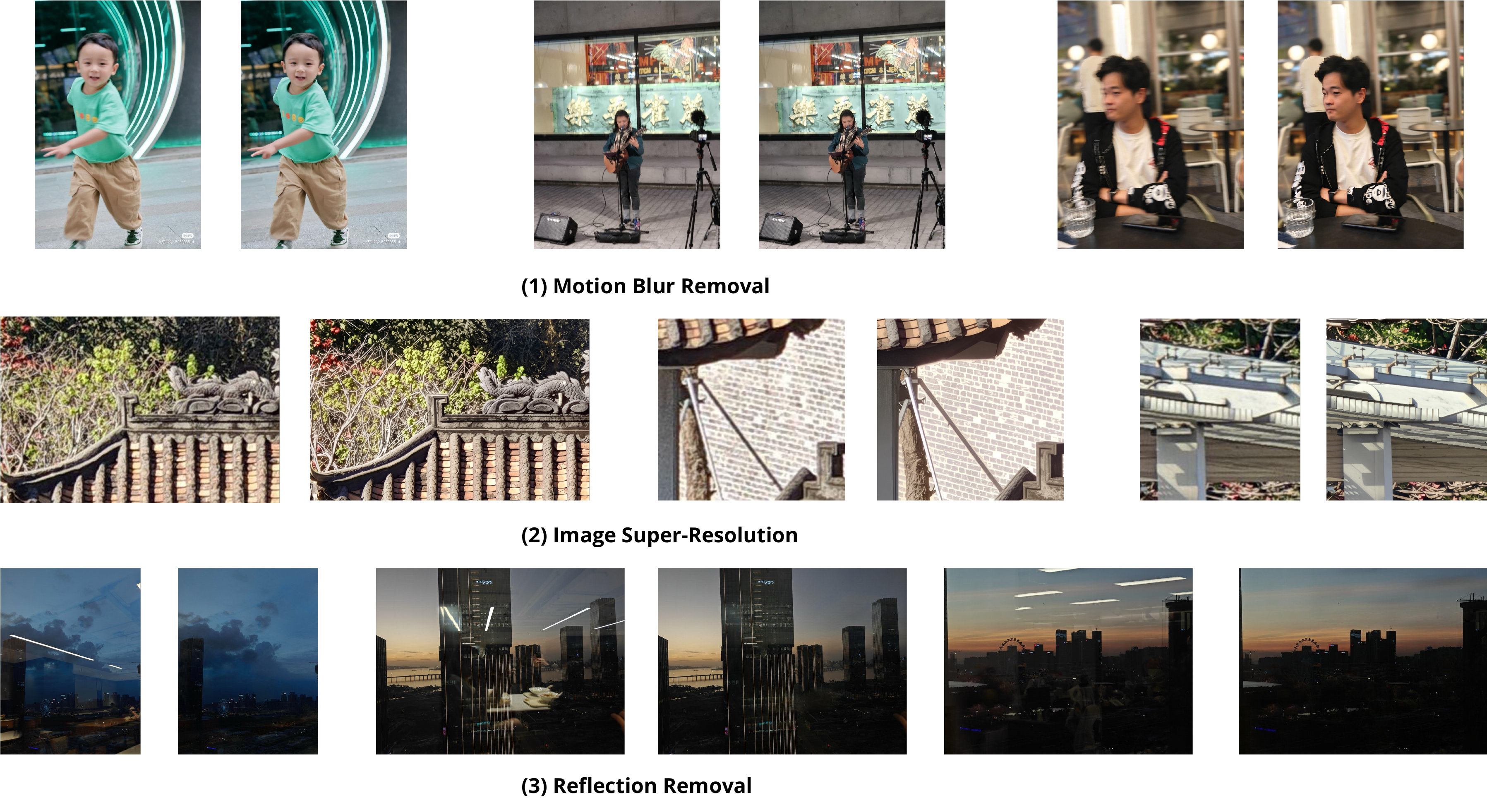}
    \caption{Restoration results for different image degradation types. The three scenarios shown are: (1) Motion blur, (2) Super-resolution degradation, (3) Reflection artifacts.}
    \label{fig:result}
\end{figure*}

The following Fig.~\ref{fig:result} presents these results across all three degradation scenarios.
The results show that our method is effective in handling each type of degradation. For super-resolution degradation, the restored images exhibit sharper details and improved clarity. In the case of reflection artifacts, our method successfully removes strong reflections and recovers the background. For motion blur, we restore image sharpness and enhance facial details, ensuring high-quality results even in challenging conditions. 
These visual examples demonstrate the robustness of our framework in tackling various real-world image degradation issues.

\section{Conclusion}
\label{sec:conclusion}

In this paper, we propose a novel framework for image degradation classification and restoration by leveraging Vision-Language Models (VLMs) and specialized restoration techniques. First, we use a VLM for zero-shot classification of degradation types, enabling a plug-and-play solution with minimal prior knowledge. For super-resolution, a diffusion model (InvSR) handles general restoration, while Real-ESRGAN enhances text regions. To address strong reflections that obscure backgrounds, we adopt an inpainting-based method using LaMa, which effectively restores affected areas for more realistic results. For motion blur, especially from human movement, we combine NAFNet for general deblurring with CodeFormer to refine facial features. Overall, this integrated and automated framework provides a scalable, high-performance solution for restoring real-world images affected by various degradations. Most importantly, these three AI-powered enhancement modules have already been deployed in OPPO AI smartphones\footnote{\url{https://www.youtube.com/watch?v=hM-ogQHHtcw&ab_channel=JieCai}}, where they process tens of thousands of user images daily.

{
\bibliographystyle{latex8}
\bibliography{reference}
}

\end{document}